\documentclass[10pt, a4paper]{article}

\usepackage{stmaryrd}

\usepackage[final]{lrec2026} 
\usepackage{times}
\usepackage{latexsym}
\usepackage[T1]{fontenc}
\usepackage[utf8]{inputenc}
\usepackage{microtype}
\usepackage{inconsolata}
\usepackage[mode=text,group-separator={,},group-minimum-digits=3,separate-uncertainty=true]{siunitx}
\usepackage[inline]{enumitem}
\usepackage{multirow}
\usepackage{rotating}
\usepackage{makecell}
\usepackage{booktabs}
\usepackage[inline]{enumitem}
\usepackage{amssymb}
\usepackage{tcolorbox}
\usepackage{cleveref}
\usepackage{xspace}
\usepackage{subcaption}
\usepackage{graphicx}

\newcommand{\modernGBERT}{ModernGBERT\xspace}
\newcommand{\Llama}{Llama\xspace}
\newcommand{\LLaMmlein}{LLäMmlein\xspace}
\newcommand{\LLaMmleinVec}{LLäMmlein2Vec\xspace}

\newcommand{\LLMVec}{LLM2Vec\xspace}
\newcommand{\QANIAH}{QA-NIAH\xspace}
\newcommand{\SuperGLEBer}{SuperGLEBer\xspace}

\newcommand{\GbertLARGE}{GBERT\textsubscript{Large}\xspace}
\newcommand{\GbertBASE}{GBERT\textsubscript{Base}\xspace}

\newcommand{\Geberta}{GeBERTa\xspace}
\newcommand{\GebertaBASE}{GeBERTa\textsubscript{Base}\xspace}
\newcommand{\GebertaLARGE}{GeBERTa\textsubscript{Large}\xspace}
\newcommand{\GebertaXLARGE}{GeBERTa\textsubscript{XLarge}\xspace}

\newcommand{\XLMRoberta}{XLM-RoBERTa\xspace}
\newcommand{\XLMRobertaBASE}{XLM-RoBERTa\textsubscript{Base}\xspace}
\newcommand{\XLMRobertaLARGE}{XLM-RoBERTa\textsubscript{Large}\xspace}
\newcommand{\XLMRobertaXLARGE}{XLM-RoBERTa\textsubscript{XLarge}\xspace}
\newcommand{\mmBERT}{mmBERT\xspace}
\newcommand{\mmBERTBase}{mmBERT-base\xspace}
\newcommand{\mmBERTSmall}{mmBERT-small\xspace}
\newcommand{\euroBERT}{EuroBERT\xspace}
\newcommand{\euroBERTsmall}{EuroBERT-210m\xspace}
\newcommand{\euroBERTmid}{EuroBERT-610m\xspace}
\newcommand{\euroBERTlarge}{EuroBERT-2b\xspace}
\newcommand{\ModernBERT}{ModernBERT\xspace}

\newcommand\emojibranding{\includegraphics[height=1.2\fontcharht\font`\B]{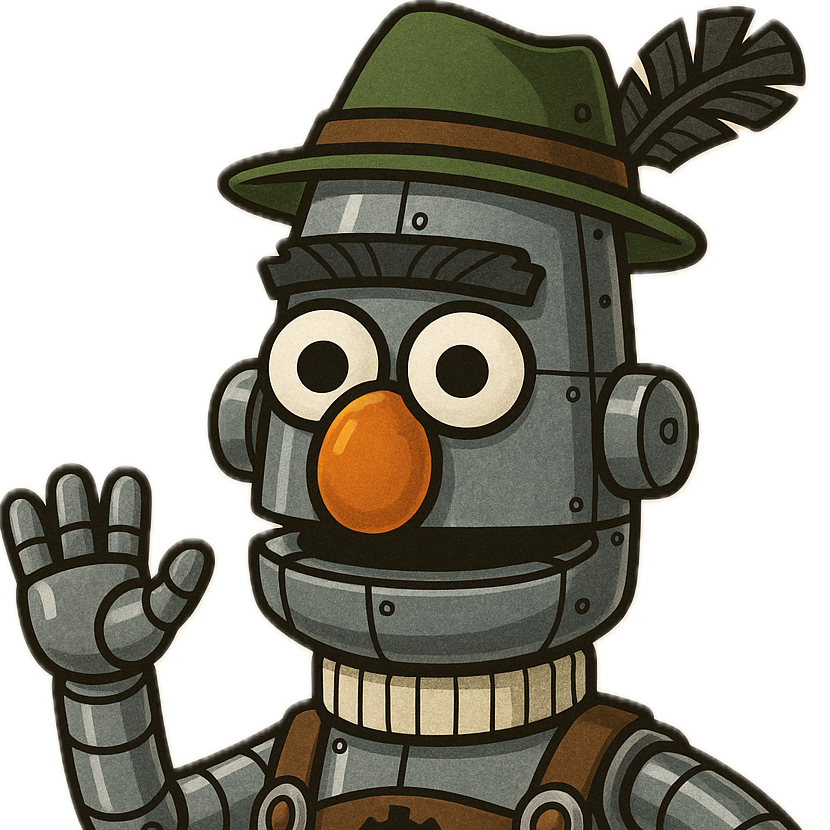}}
\newcommand\cbox[2]{\medskip\noindent\begin{tcolorbox}[width=\linewidth,colback={lightgray!40},size=small]%
\emojibranding\xspace\textbf{#1}: #2\end{tcolorbox}}
\title{New Encoders for German Trained from Scratch:\\Comparing \modernGBERT with Converted LLM2Vec Models}
\name{Julia Wunderle$^{1,*}$, Anton Ehrmanntraut$^{2,*}$, Jan Pfister$^{1}$ \\
{\bf \large Fotis Jannidis$^{2,\dagger}$, Andreas Hotho$^{1,\dagger}$}}

\address{$^1$Data Science \quad $^2$Computer Philology and History of Contemporary German Literature \\
         CAIDAS -- Center for Artificial Intelligence and Data Science \\
         JMU -- Julius-Maximilians-Universit\"at W\"urzburg \\
         {\small \{lastname\}@informatik.uni-wuerzburg.de, \{firstname.lastname\}@uni-wuerzburg.de}\\
         $^*, ^\dagger$Equal contribution}

\abstract{
Encoders remain essential for efficient German NLP and NLU scenarios despite the rise of decoder-only LLMs.
This work studies two routes to high-quality German encoders under identical data and training constraints:
\begin{enumerate*}[label=\alph*)]
    \item training from scratch and
    \item converting decoders via \LLMVec.
\end{enumerate*}
We introduce two resources:
\modernGBERT (134M, 1B), fully transparent German encoders in the \ModernBERT style, and
\LLaMmleinVec (120M, 1B, 7B), decoder-to-encoder conversions trained with masked next-token prediction,
both undergoing a context extension to \num{8192} tokens.
\\
Across \SuperGLEBer, \modernGBERT~1B sets a new state of the art (avg 0.808), surpassing \GbertLARGE (+4\%) and the seven-times larger converted 7B model (0.787).
On German MTEB after supervised fine-tuning, \modernGBERT~1B (0.551) approaches the converted 7B model (0.557).
We release all models, checkpoints, datasets, and full training records, and introduce an encoder-adapted \QANIAH evaluation.
All in all, our results provide actionable guidance:
when parameter efficiency and latency matter, from-scratch encoders dominate. When a pre-trained decoder exists and compute is a limited, conversion offers an effective alternative.\footnote{\modernGBERT and \LLaMmleinVec, including all code, data and intermediary checkpoints will be published upon acceptance under a research-only RAIL license.}
\\ \newline  }

\begin{document}
\maketitleabstract

\section{Introduction}
Encoders remain central to efficient (German) NLP and NLU where latency, memory, and cost dominate \cite{pfister-hotho-2024-supergleber}, still accounting for most of the downloads from e.g.\ Hugging Face \cite{HF_models_stats_blog_post}.
Especially for tasks like sentence similarity (i.e.\ important for RAG), mask-filling and text/token classification, encoder models are still the architecture of choice \cite{HF_models_stats_blog_post}.
Yet, the German NLP landscape lacks high-quality, modern, transparent encoders with reproducible training provenance and long-context support.
This paper contributes both resources and a controlled study addressing two practical routes to encoders under identical data and training constraints: training from scratch versus converting decoders.

We introduce to complementary model families: \modernGBERT (134M, 1B), ModernBERT-style German encoders trained from scratch, and \LLaMmleinVec (120M, 1B, 7B), encoders derived from German decoder-only models via \LLMVec.
\ModernBERT \citep{warner2024smarterbetterfasterlonger} introduced several architectural improvements for English encoders, including improved relative positional embeddings and efficient attention patterns that allow long context processing. We adapt this design to German, providing strong encoder baselines built entirely from scratch.

\begin{figure}
    \centering
    \includegraphics[width=\columnwidth]{figures/scatter_test.pdf}
    \caption{Performance on SuperGLEBer benchmark.
        $\bullet$ markers: encoders, $\blacktriangle$ markers: decoders.
        Dashed arrows: LLM2Vec conversion gains.
        Models of the same family are colored the same.
    }\label{fig:scatterplot}
\end{figure}

In parallel, to assess the practical utility and trade-offs of training encoder models from scratch, we converted the decoder-only model family \LLaMmlein into encoders using \LLMVec \citep{llm2vec}.
Specifically, to align closely with the encoder training objective, we limit the procedure to the first two \LLMVec steps:
\begin{enumerate*}
    \item enabling a bidirectional attention mask, and
    \item masked next-token prediction (MNTP).
\end{enumerate*}
Since all models share the same training datasets, this setup provides a unique foundation for systematically analyzing the relationship between different architectures and training strategies.
We extensively evaluate and compare these models during and post training via:
natural language understanding (SuperGLEBer, \citealp{pfister-hotho-2024-supergleber}), embedding performance (MTEB; \citealp{enevoldsen2025mmtebmassivemultilingualtext,muennighoff2023mtebmassivetextembedding,wehrli-etal-2023-german}), long-context understanding (new Question Answering Needle-in-a-Haystack (\QANIAH)) and an efficiency suite reflecting variable-length inference.
Our key contributions are:
\begin{itemize}[leftmargin=*]
    \itemsep0em
    \item We introduce a \modernGBERT family, which achieves new state-of-the-art performance on \SuperGLEBer and the German MTEB.
    \item To enable comparisons between training strategies, we also introduce a decoder-turned-encoder \LLaMmleinVec family, based on the same training dataset.
    \item We find that dedicated encoders trained from scratch consistently outperform converted decoders of similar size.
    \item Our newly introduced \QANIAH benchmark confirms strong long-context understanding of our models, validating the effectiveness for extended input sequences.
    \item To enable further research we release all (intermediary) resources including all model checkpoints, data point tracking, code and data.
\end{itemize}
\cbox{Note}{Throughout the paper, we highlight interesting findings and insights we gained during the process in little boxes like this one.}

\section{Datasets}
\subsection{Pre-Training Dataset}\label{sec:pretrain-dataset}
We pre-trained \modernGBERT on the same data as \LLaMmlein decoder models \citep{llammlein}, using the open-source RedPajamaV2 dataset \cite{NEURIPS2024_d3449733}
.
This dataset comprises German CommonCrawl snapshots from 2014 to 2023.
As we intend to keep datasets constant between \modernGBERT and \LLaMmlein, we follow \LLaMmlein's data pipeline and select the higher quality document-level deduplicated ``head'' and ``middle'' partitions, excluding the lower quality ``tail'' partition.
For our 134M model, we only selected the head partition.
Our processing pipeline mirrors \citet{llammlein}: we first perform paragraph-level deduplication using a Bloom filter to remove redundant content such as GDPR notices and web boilerplate, enhancing data diversity.
Afterward, we apply a token-to-word ratio filter to further improve text quality.
The final dataset is about \SI{6}{\tera\byte}, corresponding to approximately 1.27T tokens using a \GbertLARGE tokenizer.

\subsection{Context Extension Datasets}\label{sec:contextext-dataset}
ModernBERT enhances its context capacity from \num{1024} to \num{8192} by fine-tuning in two phases:
on an 250B-token subset of \num{8192}-token sequences from the original pre-training data (\textit{ext1}), followed by 50B-token high-quality dataset with mixed sequence lengths (\textit{ext2}) \cite{gao2025trainlongcontextlanguagemodels}.

Following this setup, we construct two German datasets.
For \textit{ext1}, we analogously subsample long sequences from our pre-training dataset, resulting in ``\textit{LONG-Head}'' from the head partition for our 134M model, and ``\textit{LONG-Head/Middle}'' from the head and middle partition for our 1B model.

For the high-quality dataset used for \textit{ext2}, we selected the German portion of Fineweb2 as basis \citep{penedo2024fineweb-2}
.
To match the original distribution, we take a randomized sample of Fineweb2, and add a separate Fineweb2 sample, selecting long documents with $\num{>=8192}$ tokens, splitting them into sequences of about \num{8192} tokens (``Fineweb2-long'').
Additional long documents are drawn from the 2023 German Wikipedia
and the 2022 OpenLegalData dump,
.
The resulting dataset contains 14.4B tokens, summarized in \Cref{tab:context-dataset}.

\begin{table}
    \centering
    \resizebox{\columnwidth}{!}{%
        \begin{tabular}{llrrr}
            \toprule
             & \textbf{Dataset}       & \textbf{\# Tokens} & \textbf{\# Sequences} & \textbf{\makecell{Median \\Length}}\\
            \midrule
            \multirow{2}{*}{\rotatebox{90}{\textit{ext1}}} & \textit{LONG-Head}             & \num{52}B      & \num{6813019}     & \num{7755}                  \\
                                                 & \textit{LONG-Head/Middle}           & \num{90}B      & \num{11785941}    & \num{8013}                  \\
            \midrule
            \multirow{5}{*}{\rotatebox{90}{\textit{ext2}}} & \textit{High Quality}                 & \num{14.4}B    & \num{43191271}    & \num{199}                   \\
                                                 & $\rotatebox[origin=c]{180}{$\Lsh$}$ Fineweb2       & \num{7640}M    & \num{42319173}    & \num{194}                   \\
                                                 & $\rotatebox[origin=c]{180}{$\Lsh$}$ Fineweb2-long  & \num{6211}M    & \num{799296}      & \num{7902}                  \\
                                                 & $\rotatebox[origin=c]{180}{$\Lsh$}$ OpenLegalData  & \num{407}M     & \num{53798}       & \num{7583}                  \\
                                                 & $\rotatebox[origin=c]{180}{$\Lsh$}$ Wikipedia      & \num{143}M     & \num{19004}       & \num{7515}                  \\
            \bottomrule
        \end{tabular}
     }%
    \caption{Composition of post-training datasets for context extension and decoder conversion.}\label{tab:context-dataset}
\end{table}

\section{Methodology \protect\footnote{All training settings and parameters will be detailed in the camera-ready version.}}
\subsection{\modernGBERT}\label{sec:moderngbert}
We adapt the \ModernBERT architecture and training strategy for German.
While \modernGBERT 134M matches the base \ModernBERT model size (22 layers, 768 hidden units, but 16M fewer parameters due to a smaller vocabulary size), we create \modernGBERT 1B with 28 layers and a hidden size of \num{2048}.

\paragraph{Pre-/Post-Training Strategy}
Both models are pre-trained using masked language modeling (MLM) with no next-sentence-prediction, a 30\% masking rate, and sequences up to \num{1024} tokens (\num{10000} RoPE theta).
\modernGBERT 134M is trained on the head partition (0.47T tokens), as downstream evaluation indicated early saturation, while \modernGBERT 1B is trained on both the head and middle partitions of our pre-training corpus (\Cref{sec:pretrain-dataset}), totaling 1.27T tokens.

After standard MLM pre-training, we proceed with the two context extension phases:
During \textit{ext1}, the RoPE theta is raised to \num{160000}, and models are trained on the \textit{LONG-Head} (134M) or \textit{LONG-Head/Middle} (1B) datasets (see \Cref{sec:contextext-dataset}).
In the \textit{ext2} phase, both models are trained on the \textit{High Quality} dataset.

\paragraph{Tokenization}
For tokenization, we use the original BERT-style tokenizer from \GbertLARGE, resulting in a \num{31168}-word embedding layer.
While \LLaMmlein \citep{llammlein} provides a dedicated German BPE tokenizer, our preliminary ablations using this tokenizer consistently showed degraded downstream performance.
Consequently, we retained the \GbertLARGE{} tokenizer.

\paragraph{Training Progress Tracking}
Throughout the training we save, evaluate and release all checkpoints to support further research.
Inspired by Pythia \citep{pmlr-v202-biderman23a} we provide full training transparency by logging and releasing the order of data points seen during training; thus, all checkpoints can be linked with the exact data points seen up to that checkpoint.

\subsection{\LLMVec: Turning Decoders to Encoders}\label{sec:methodology_llm2vec}
\citet{llm2vec} proposes a method to convert decoder-only LLMs into text encoders through the following steps:
First, the causal attention mask is replaced with a full attention mask, enabling bidirectional attention across tokens.
Second, the model is trained using a masked next token prediction (MNTP) objective.
Third, regular \LLMVec includes a unsupervised contrastive learning (SimCSE) step, improving embedding quality by maximizing agreement between differently dropped-out versions of the same input.
However, we intend to remain as close as possible to the training objectives of \modernGBERT to allow fair and direct comparisons.
Therefore, we limit the procedure to the first two steps of LLM2Vec, specifically employing the MNTP objective, which is most closely aligned with MLM.

\paragraph{Pre-/Post-Training Strategy}
We train all three \LLaMmlein models (120M, 1B, 7B) using the same two context extension datasets as employed by \modernGBERT's context extensions (\textit{ext1} and \textit{ext2}) (\Cref{sec:contextext-dataset}).
Specifically, the 120M model mirrors \modernGBERT 134M, being trained on the \textit{LONG-Head} (\textit{ext1}) and \textit{High Quality} (\textit{ext2}) datasets, while the 1B and 7B models follow \modernGBERT 1B, using \textit{LONG-Head/Middle} (\textit{ext1}) and the \textit{High Quality} dataset (\textit{ext2}).
Notably, MNTP training is applied separately to each dataset, resulting in two adapter modules for each model.
Additionally, all adapters were trained on the full corresponding datasets\footnote{except for the \LLaMmleinVec 7B trained on the \textit{LONG-Head/Middle} model, which we trained on 64 nodes with 4 H200 each for 14 hours, before stopping the training due to compute constraints.}, although models were able to achieve comparable results during earlier training stages. This indicates, that compute time could have been drastically reduced.

Further aligning with \modernGBERT we combined MNTP with context extension — a setup that, to our knowledge, has not yet been widely explored.
Therefore, we also extended the model sequence length to \num{8192} tokens and increased RoPE theta to \num{160000}.
We evaluate both individual adapters (\textit{ext1} \& \textit{ext2}) as well as a merged model (\textit{ext1+2}) that combines both

\section{Evaluation Setup}
\subsection{\SuperGLEBer}\label{sec:supergleber}
We assess our final models using the German \SuperGLEBer benchmark \cite{pfister-hotho-2024-supergleber}, which includes 29 tasks across text classification, sequence tagging, question answering, and sentence similarity.
These tasks cover diverse domains such as news, legal texts, and consumer reviews.
For each task, models are fine-tuned with QLoRA \cite{dettmers2023qloraefficientfinetuningquantized} by default, or LoRA as fallback.
In addition to evaluating final checkpoints, we follow \LLaMmlein~\cite{llammlein} and evaluate intermediate checkpoints of \modernGBERT as well as \LLaMmleinVec on the same representative \SuperGLEBer subset consisting of:
the classification tasks NLI~\cite{conneau-etal-2018-xnli}, FactClaiming Comments~\cite{risch-etal-2021-overview}, DB Aspect~\cite{Wojatzki2017GermEval2S}, and WebCAGe~\cite{henrich-etal-2012-webcage}, the sequence tagging task EuroParl~\cite{Faruqui2010TrainingAE}, and the sentence similarity task PAWSX~\cite{liang-etal-2020-xglue}.

\subsection{Massive Text Embedding Benchmark}
We further evaluate the models on the German subset of the Massive Text Embedding Benchmark \textit{MTEB(deu,v1)} \citep{enevoldsen2025mmtebmassivemultilingualtext}, comprising 19 tasks.
In addition to text pair classification and semantic textual similarity, which are already covered by the \SuperGLEBer benchmark, MTEB includes clustering \cite{wehrli-etal-2023-german}, as well as reranking and retrieval tasks.
These latter tasks provide a more comprehensive assessment of general-purpose sentence embeddings, focusing on the models' ability to produce robust semantic representations.

To adapt the base models for embedding tasks, we fine-tune them using the Sentence-Transformer framework \citep{reimers-gurevych-2019-sentence} in a supervised setup.
Fine-tuning employs \num{10000} samples from the German portion of the machine-translated multilingual mMARCO passage ranking dataset \citep{bonifacio2022mmarcomultilingualversionms}, maximizing similarity between query and positive passages, while minimizing similarity to negative passages.
Sentence embeddings are obtained by mean pooling over the final token representations.
We use InfoNCE loss with a batch size of 128 and a learning rate of \num{5e-5}.
We apply QLoRA for efficient training (falling back to LoRA for the GBERT family, where quantization is not supported).

\subsection{Long-Context Understanding}
Evaluating long-context capabilities in German is hindered by the scarcity of native high-quality datasets, with translations from English often introducing artifacts.
To address this, we construct a new Question-Answering Needle-In-a-Haystack (\QANIAH) evaluation \citep{ivgi-etal-2023-efficient,hsieh2024rulerwhatsrealcontext} based on the human-annotated GermanQuAD dataset \citep{moller-etal-2021-germanquad}.
To our knowledge, this is the first \QANIAH evaluation specifically adapted for encoder models.
The goal in \QANIAH is to extract an answer span from a long document.
We adapt GermanQuAD to a \QANIAH setup as follows:
for each question-paragraph (``needle'') pair, we sample up to 3 distractor paragraphs and shuffle them with the needle, forming a ``haystack'' document of up to \num{1024} tokens.
The answer always appears only in the needle paragraph.
For evaluation, up to 20 distractors are included to test generalization to longer contexts, yielding documents up to \num{8192} tokens.
This results in \num{11518} training and \num{2204} test question--haystack pairs.
Models were again fine-tuned using QLoRA falling back to LoRA, where quantization is not supported.
We will release the \QANIAH generation code and splits to facilitate reuse and extension.

\section{Evaluation Results}
The tables in this paper present only a summary of our results.
Extensive benchmarking results, analyses, and all hyperparameters of all models will be provided in the camera-ready version.

\begin{figure*}
    \centering
    \begin{subfigure}{\textwidth}
        \centering
        \includegraphics[width=\textwidth]{figures/training_log_1B_PAWSX.pdf}
    \end{subfigure}
    \hfill
    \begin{subfigure}{\textwidth}
        \centering
        \includegraphics[width=\textwidth]{figures/training_log_llm2vec_1b_with_pawsx.pdf}
    \end{subfigure}
    \caption{Intermediate checkpoint evaluation. The solid black line shows the mean of six SuperGLEBer tasks (NLI, FactClaiming Comments, DB Aspect, WebCAGe, EuroParl, PAWSX Similarity). The top figure shows \modernGBERT 1B across pre-training and two context extension phases, with box plots representing all 29 SuperGLEBer tasks. For simplicity, no significant improvements between checkpoint pairs are marked with brackets (Wilcoxon signed-rank test). All other pairs of box plots show significant improvements of at least p < 0.01. The bottom figure shows \LLaMmlein 1B after \LLMVec conversion, including the starting point in (green) and the average score after switching the mask (red). Checkpoints on \textit{ext2} are shown alone and merged with the last \textit{ext1} checkpoint.}
    \label{fig:training}
\end{figure*}

\subsection{Intermediate Model Evaluation}\label{sec:pre-train-evaluation}
\paragraph{\modernGBERT Training}

To track the pre-training progress, we evaluated intermediate checkpoints on the \SuperGLEBer benchmark (see \Cref{sec:supergleber}). \Cref{fig:training} exemplary shows the results for the \modernGBERT 1B model in the top row.
To quantify trends, we first assessed the full \SuperGLEBer suite on five 1B and respective three 134M equally spaced \modernGBERT checkpoints and performed Wilcoxon signed-rank tests.
The 134M model plateaued after 72B tokens (15\% of data), with no further significant gains.
In contrast, the 1B model achieved significant improvements after the same amount of data (p $<$ 0.0001) and on the middle partition (p $<$ 0.00052), before plateauing after 864B tokens (67\% of data), with only minor gains thereafter (0.777 $\rightarrow$ 0.791)

To gain more insights, we tracked six representative \SuperGLEBer subtasks at each checkpoint.
For the 134M model, only PAWSX showed a positive Spearman rank correlation with training duration (r = 0.655; p $<$ 0.003), whereas for 1B, all tasks except EuroParl improved significantly (r $>$ 0.57; p < 0.00014), and complex tasks like PAWSX continued to show modest gains even after the aggregate score stabilized (gray in \Cref{fig:training}).

These saturation patterns, including per-task trends and overall performance plateaus, are consistent with findings by \citet{llammlein} for decoder models, and by \citet{antoun2024camembert20smarterfrench} for the French ModernBERT variant ModernCamemBERT (136M parameters).
Our results confirm that while small ModernBERT models saturate quickly, larger models continue to benefit from additional data.
Extrapolating from this observed scaling behaviour between \modernGBERT 134M and 1B,
we hypothesize that a larger 7B encoder could leverage extensive monolingual corpora to achieve performance beyond \modernGBERT 1B.

\cbox{Confirmation}{Our findings corroborate \citet{antoun2024camembert20smarterfrench} and \citet{llammlein}: small \modernGBERT models reach saturation early, while scaling model and dataset size enables improvements.}

\paragraph{\LLaMmleinVec Conversion}
To remain as consistent as possible with the \modernGBERT training procedure, we trained the \LLaMmleinVec conversions on the same post-training datasets.
To assess whether training on the full dataset was necessary, we evaluated multiple checkpoints throughout the entire training process on the six \SuperGLEBer tasks.
In the bottom half of \Cref{fig:training} we illustrate the average SuperGLEBer performance alongside the of the standard decoder models (blue), as well as the performance of decoders with a bidirectional attention mask (purple).
Interestingly, for encoder-typical tasks such as sentence similarity (PAWSX), the performance of both the 120M, 1B and 7B \LLaMmlein models increased even without explicit bidirectional continued pre-training, showing gains from bidirectional fine-tuning alone (120M: 0.477 $\rightarrow$ 0.516; 1B: 0.548 $\rightarrow$ 0.580; 7B: 0.524 $\rightarrow$ 0.660).

\cbox{Observation}{Switching the decoder attention mask to bidirectional already yields improvements on encoder-specific tasks without continued pre-training but bidirectional fine-tuning.}

Regarding the increases in average performance, the scores remained largely consistent over time, without showing significant improvements. This suggests that training could have been stopped earlier, thus reducing the overall training time, emphasizing that converting a decoder is substantially faster than training a model from scratch.

\subsection{Final Model Evaluation}
\begin{table}[t]
    \centering
    \small
    \resizebox{\columnwidth}{!}{%
        \begin{tabular}{lcccccc}
            \toprule
            \textbf{Model}    & \textbf{Size}           & \textbf{Avg.}   & \textbf{Class.} & \textbf{NER}   & \textbf{PAWSX} & \textbf{QA}    \\
            \midrule
          \GbertBASE     & 111M                & 0.718          & 0.723           & 0.786          & 0.561          & 0.803          \\
            \GbertLARGE     &  337M             & 0.768          & 0.785           & 0.799          & 0.654          & 0.832          \\[1ex]
            \GebertaBASE &   139M                 & 0.716          & 0.715           & 0.778          & 0.559          & 0.813          \\
            \GebertaLARGE   &     406M            & 0.749          & 0.743           & 0.791          & 0.619          & 0.844          \\
            \GebertaXLARGE  &   887M              & 0.767          & 0.770           & 0.807          & 0.643          & 0.848          \\[1ex]
            \XLMRobertaBASE   &   279M                 & 0.689          & 0.693           & 0.754          & 0.505          & 0.802          \\

            \XLMRobertaLARGE  &    561M      & 0.730          & 0.714           & 0.787          & 0.583          & 0.837          \\
            \XLMRobertaXLARGE &    3.48B           & 0.758          & 0.750           & 0.802          & 0.656          & 0.822          \\

            \midrule
                        \mmBERTSmall&  142M & 0.748 & 0.759 & 0.800& 0.606& 0.828\\

            \mmBERTBase &  309M & 0.758 & 0.757 & 0.804 & 0.622& 0.849\\[1ex]
                        \euroBERTsmall & 212M  & 0.706& 0.659& 0.698& 0.607& 0.859\\
                        \euroBERTmid & 609M  & 0.721& 0.633& 0.750& 0.633& 0.869\\
                        \euroBERTlarge & 2.11B & 0.731& 0.633& 0.764 &0.655 & 0.873\\

            \midrule
            \LLaMmlein &  120M                & 0.676          & 0.702           & 0.712          & 0.477          & 0.812          \\

            \LLaMmleinVec (\textit{ext1}) &   120M             & 0.684          & 0.703           & 0.741          & 0.472          & 0.819          \\[1ex]
            \LLaMmlein &   1B                  & 0.733          & 0.781           & 0.773          & 0.548          & 0.828          \\

            \LLaMmleinVec (\textit{ext1}) &  1B               & 0.762          & 0.776           & 0.812          & 0.615          & 0.843          \\[1ex]
            \LLaMmlein &  7B                  & 0.747          & 0.810           & 0.805          & 0.524          & 0.851
            \\
            \LLaMmleinVec (\textit{ext1}) &  7B               & 0.787          & 0.799           & 0.838          &
            \bfseries 0.670                & 0.842                                                                               \\
            \midrule
            \modernGBERT &  134M & 0.730          & 0.716           & 0.782          & 0.589          & 0.833          \\

            \modernGBERT (\textit{ext1+2}) &  134M              & 0.749          & 0.735           & 0.805          & 0.612          & 0.836          \\[1ex]
            \modernGBERT &  1B  & 0.800          & 0.806           & 0.839          & 0.681          & 0.874          \\

            \modernGBERT (\textit{ext1+2}) &  1B                & \textbf{0.808} & \textbf{0.812}  & \textbf{0.845} & 0.699          & \textbf{0.876} \\

            \bottomrule
        \end{tabular}
    }
    \caption{Performance comparison on \SuperGLEBer benchmark. For \modernGBERT we provide results before and after the context extension.}
    \label{tab:supergleb_small}
\end{table}

\paragraph{Natural Language Understanding}
We evaluate all final models on the full \SuperGLEBer benchmark.
\Cref{tab:supergleb_small} reports average performances across a subset of our models.
In particular, we compare our models to established encoders:
GBERT \citep{chan-etal-2020-germans}, \Geberta{} \citep{dada-etal-2023-impact}, \XLMRoberta{} \citep{conneau-etal-2020-unsupervised,goyal-etal-2021-larger}, \mmBERT \cite{marone2025mmbertmodernmultilingualencoder} and \euroBERT \citep{boizard2025eurobert}.

Our \modernGBERT consistently outperforms comparable and larger models.
The 134M variant achieves an average score of 0.749, surpassing all similar-sized baselines, as well as \LLaMmlein 1B (0.733), \euroBERTlarge (0.731) and \LLaMmlein 7B (0.747).
The \modernGBERT 1B variant achieves a new state-of-the-art average score across the entire \SuperGLEBer of 0.808, outperforming \GbertLARGE (0.768) by 4\% and beating the seven times larger \LLaMmleinVec 7B (0.787).
It leads in three of four evaluation categories, including classification (0.812), NER (0.845), and QA (0.876).
Only on sentence similarity (0.699), \LLaMmleinVec 7B achieves better results.
\modernGBERT scales well, with performance improving for larger model sizes, again suggesting that scaling ModernBERT-style encoders can leverage large monolingual corpora effectively.
In the \SuperGLEBer setting, adding context extension improved \modernGBERT's average by 1.9\% for the 134M model (from ~0.730 to 0.749) and by 0.8\% for the 1B variant (from ~0.800 to 0.808).
However, large improvements were not expected, as \SuperGLEBer tasks do not make use of long contexts.

Adaptation via \LLMVec yields consistent gains across models.
Our first \LLMVec tuning (on \textit{ext1}, \Cref{sec:contextext-dataset}) showed the most prominent positive effect, while the second fine-tune using the \textit{ext2} datasets showed only marginal increase, or even a decrease in performance.
The same holds for a mixture of the two \LLMVec adapters (\textit{ext1+2}). Therefore, we report only \textit{ext1} results in this and all subsequent tables for simplicity.
The \LLaMmleinVec 7B achieves the strongest results among the \LLMVec models (0.787).
Conversion of \LLaMmlein 120M, 1B, 7B improved the average score by 0.8\%, 2.9\%, and 4.0\% respectively.
This effect is especially pronounced in PAWSX, with scores increasing by up to 14.6\% for \LLaMmlein 7B and 6.7\% for \LLaMmlein 1B.
\cbox{Observation}{\LLMVec yields the best improvement on similarity-related tasks.}

Comparing the \LLaMmleinVec with the \modernGBERT family, we find that on similarly sized models, \modernGBERT always outperforms the transformed decoders by a large margin.
Only the much larger \LLaMmleinVec 7B approaches the performance of \modernGBERT 1B.
This systematic comparison using identical datasets provides the first comprehensive analysis of MLM vs. LLM2Vec for encoder development.
\cbox{Observation}{With similar data and model sizes, training encoders from scratch outperforms our converted models.}

\begin{table}
    \centering
    \small
    \resizebox{\columnwidth}{!}{%
        \begin{tabular}{lccccc}
            \toprule
            \textbf{Model}          & \textbf{Size}                        & \textbf{Avg.}     & \textbf{Clstr} & \textbf{ReRnk} & \textbf{Retr} \ \\
            \midrule
            \GbertBASE$^{\phantom{f}}$  & 111M           & 0.360           & 0.274               & 0.118              & 0.226                \\
            \GbertBASE${}^f$                 &   111M & 0.500           & 0.318               & 0.374              & 0.461                \\[1ex]
            \GbertLARGE$^{\phantom{f}}$        & 337M       & 0.412           & 0.336               & 0.206              & 0.297                \\
            \GbertLARGE${}^f$                        & 337M & 0.521           & 0.334               & 0.389              & 0.493                \\[1ex]

            \GebertaBASE$^{\phantom{f}}$ & 139M & 0.382& 0.312 & 0.174& 0.213\\
            \GebertaBASE${}^f$ & 139M & 0.493& 0.318 & 0.374& 0.430\\[1ex]

            \GebertaLARGE$^{\phantom{f}}$  & 406M & 0.397& 0.287 & 0.223 & 0.274\\
            \GebertaLARGE${}^f$ & 406M & 0.494& 0.311 & 0.374& 0.432\\[1ex]

            \GebertaXLARGE$^{\phantom{f}}$ &887M  & 0.325& 0.278 & 0.108& 0.058\\                  \GebertaXLARGE${}^f$ & 887M & 0.521& 0.323 & 0.414& 0.462\\[1ex]

            \XLMRobertaBASE$^{\phantom{f}}$ & 279M           & 0.248           & 0.173               & 0.024              & 0.008                \\
            \XLMRobertaBASE${}^f$                 & 279M   & 0.403           & 0.247               & 0.247              & 0.299                \\ [1ex]

            \XLMRobertaLARGE$^{\phantom{f}}$  & 561M        & 0.264           & 0.172               & 0.048              & 0.026                \\
            \XLMRobertaLARGE${}^f$                  & 561M  & 0.460           & 0.259               & 0.343              & 0.416                \\[1ex]

            \XLMRobertaXLARGE$^{\phantom{f}}$      & 3.48B   & 0.301           & 0.225               & 0.090              & 0.142                \\
            \XLMRobertaXLARGE${}^f$                  & 3.48B   & 0.479           & 0.342               & 0.362              & 0.407                \\

            \midrule
            \mmBERTSmall & 142M & 0.289 & 0.163 & 0.091& 0.075\\
            \mmBERTSmall${}^f$  & 142M& 0.430 &0.167 & 0.390 & 0.359 \\[1ex]

            \mmBERTBase & 309M& 0.318 & 0.237 &0.092 & 0.071 \\
                        \mmBERTBase${}^f$ & 309M & 0.475 & 0.214 & 0.428 & 0.406 \\[1ex]

                        \euroBERTsmall & 212M & 0.293 & 0.240 & 0.070 & 0.118\\
                                                \euroBERTsmall${}^f$ & 212M & 0.431 & 0.189 & 0.404 & 0.371 \\[1ex]

                        \euroBERTmid & 609M& 0.299 & 0.232 & 0.093 & 0.134\\
                                                \euroBERTmid${}^f$ & 609M & 0.419& 0.288& 0.263 & 0.352\\[1ex]

                        \euroBERTlarge &2.11B & 0.230  & 0.186 & 0.060 & 0.065\\
                                                \euroBERTlarge${}^f$ &2.11B &  0.452& 0.191 &0.460 & 0.420\\

            \midrule
            \LLaMmleinVec (\textit{ext1})$^{\phantom{f}}$   & 120M     & 0.315           & 0.261               & 0.139              & 0.224                \\
            \LLaMmleinVec  (\textit{ext1})${}^f$    &      120M         & 0.471           & 0.308               & 0.325              & 0.425                \\[1ex]

            \LLaMmleinVec (\textit{ext1})$^{\phantom{f}}$   & 1B      & 0.399           & 0.308               & 0.183              & 0.276                \\
            \LLaMmleinVec (\textit{ext1})${}^f$    &      1B           & 0.540           & \bfseries 0.343     & 0.433              & 0.511                \\[1ex]

            \LLaMmleinVec  (\textit{ext1})$^{\phantom{f}}$   & 7B       & 0.376           & 0.249               & 0.169              & 0.266                \\
            \LLaMmleinVec  (\textit{ext1})${}^f$  &       7B            & \bfseries 0.557 & 0.339               & \bfseries 0.477    & \bfseries 0.522      \\

            \midrule

            \modernGBERT$^{\phantom{f}}$ &  134M & 0.383          & 0.293               & 0.139              & 0.241                \\
            \modernGBERT${}^{f}$   &   134M    & 0.485           & 0.303               & 0.364              & 0.432                \\[1ex]

            \modernGBERT (\textit{ext1+2}) $^{\phantom{f}}$ &       134M  & 0.376           & 0.296               & 0.120              & 0.213                \\
            \modernGBERT (\textit{ext1+2}) ${}^f$             &    134M   & 0.501           & 0.312               & 0.404              & 0.446                \\[1ex]

            \modernGBERT $^{\phantom{f}}$ & 1B  & 0.374           & 0.318               & 0.097              & 0.199                \\
            \modernGBERT  ${}^{ f}$         &  1B & 0.549           & 0.339               & 0.463              & 0.511                \\[1ex]

            \modernGBERT  (\textit{ext1+2}) $^{\phantom{f}}$    &  1B      & 0.366           & 0.307               & 0.088              & 0.191                \\
            \modernGBERT  (\textit{ext1+2}) ${}^f$           &     1B      & 0.551           & 0.338               & 0.459              & 0.512                \\

            \bottomrule
        \end{tabular}
    }
    \caption{Performance comparison on MTEB. ``Avg.'' refers to the average over all six task groups, not only the ones shown here. $f$ marks the variants with additional training.}
    \label{tab:mteb_small}
\end{table}

\paragraph{Text Embedding}
We evaluate models on the MTEB benchmark, which covers six task categories:
classification, pair classification, clustering, reranking, retrieval, and short text similarity (STS) tasks.
A summary of results is visualized in \Cref{tab:mteb_small}.
In general, supervised fine-tuning on mMARCO yields consistent improvements across all model types.
While classification performance sometimes declines, substantial gains can be observed in other areas:
25\% on average for reranking, 26\% for retrieval and 25\% for STS.
\cbox{Observation}{Fine-tuning yields the largest gains in reranking, retrieval, and STS tasks, rather than for classification and clustering.}
The best overall average performance is achieved by the fine-tuned \LLaMmleinVec 7B (0.557), closely followed by the fine-tuned \modernGBERT 1B (0.551), despite the latter being significantly smaller.
\LLaMmleinVec models generally show strong performance after fine-tuning, particularly when trained with the extension dataset of the first phase (\textit{ext1}).
Using the second extension phase (\textit{ext2}) or combining both adapters into the base model (\textit{ext1+2}) again harms the performance.
Interestingly, the latter shows the largest fine-tuning gains among the three variants.
The \modernGBERT models perform competitively to similarly sized models.
Before fine-tuning, \modernGBERT 1B (avg.\@ 0.366) already outperforms most encoder-only models, such as \GebertaXLARGE{} (0.325), \XLMRobertaXLARGE{} (0.301), and \euroBERTlarge (0.230) but not \GbertLARGE{} (0.412).
However, after fine-tuning, it demonstrates clear superiority among native encoder-only models by at least 3\% on the average score.
As with our observations on the SuperGLEBer benchmark, \modernGBERT's context extension did not show significant improvements here.
Comparing \modernGBERT and \LLaMmleinVec, we find that before fine-tuning, the \LLaMmleinVec 1B and 7B models produce better representations than \modernGBERT 1B.
However, after fine-tuning, \modernGBERT 1B surpasses the 1B variant of \LLaMmleinVec on average and closely aligns with the larger 7B model.

\paragraph{Long-Context Understanding}

\Cref{tab:niah-small} reports results on our new German \QANIAH benchmark .
We evaluate subsets of short (\num{<1024}), medium (\num{1024} to \num{4095}), and long (\num{4096} to \num{8192}) sequences, focusing on LLMs supporting up to \num{8192} tokens:
\modernGBERT, the encoder-converted \LLaMmleinVec, as well as their original decoder counterparts.
Notably, \LLaMmlein models were pre-trained with a maximum context of \num{2048} tokens and only \LLaMmleinVec was post-trained with a maximum context length of \num{8192}.
\modernGBERT~1B demonstrates strong long-context performance across all lengths, outperforming all encoders. The first extension phase during training approximately tripled accuracy, while the final \textit{High Quality} extension slightly reduced performance, especially for the 134M variant.
Regarding \LLMVec, a sufficiently long conversion improved long-context understanding.
Conversion of \LLaMmlein 120M and 1B decoders (with native context length of \num{2048}) improved accuracy by factor 1.3 resp.\@ 2, both not as pronounced in comparison to the \modernGBERT encoders.
For \LLaMmleinVec 7B however (with \LLMVec training on approximately half of our \textit{ext1} dataset), it decreased by 51\%, with no correct answers on haystacks of \num{>4096} tokens.
Given the intensive compute requirements, we did not explore further optimizations regarding context extension of the \LLaMmleinVec 7B model.
\cbox{Observation}{On small training datasets, \LLMVec tuning limits the understanding of long-context samples.}

\begin{table}
        \centering
        \small
            \resizebox{\columnwidth}{!}{%

        \begin{tabular}{l@{}rcccc}
                \toprule
                \textbf{Model}                          &  \textbf{Size} &
                {\makecell[b]{\num{<1024} tok.}} &
                {\makecell[b]{\num{1024} to                                                                                       \\\num{4095} tok.}} &
                {\makecell[b]{\num{4096} to                                                                                       \\\num{8192} tok.}} &  \textbf{Avg.} \\
                \midrule
                \LLaMmlein                       & 120M     & 0.286           & 0.124           & 0.049             & 0.091      \\
                \LLaMmlein                       & 1B       & 0.517           & 0.230           & 0.088                  & 0.165\\
                \LLaMmlein                       & 7B       & 0.529           & 0.310           & 0.122                   & 0.216\\
                \midrule
                \LLaMmleinVec (\textit{ext1})             & 120M     & 0.315           & 0.206           & 0.044                   & 0.120\\
                \LLaMmleinVec (\textit{ext1})             & 1B       & 0.588           & 0.448           & 0.232                  & 0.333\\
                \LLaMmleinVec (\textit{ext1})             & 7B       & 0.597           & 0.207           & 0.000                & 0.111\\
                \midrule
                \modernGBERT                     & 134M     & 0.552           & 0.168           & 0.013                 & 0.105\\
                \modernGBERT (\textit{ext1})              & 134M     & 0.536           & 0.410           & 0.238                 & 0.323\\
                \modernGBERT (\textit{ext1+2})            & 134M     & 0.540           & 0.393           & 0.201               & 9.296 \\[1ex]
                \modernGBERT                     & 1B       & 0.556           & 0.233           & 0.023              & 0.136\\
                \modernGBERT (\textit{ext1})               & 1B       & \bfseries 0.617 & 0.506           & \textbf{0.406} & 0.457 \\
                \modernGBERT  (\textit{ext1+2})           & 1B                 & \bfseries 0.526 & 0.383          & 0.451          & 0.451 \\
                \bottomrule
        \end{tabular}
        }
        \caption{\QANIAH results, wiht Exact Match metric. All tokens are counted per the model's respective tokenizer.}\label{tab:niah-small}
\end{table}

\subsection{Inference Efficiency}
\begin{table}
   \small
   \centering
       \resizebox{\columnwidth}{!}{%

   \begin{tabular}{l@{} r S[table-format=3.2(2),separate-uncertainty] S[table-format=3.2(2),separate-uncertainty] S[table-format=3.2(2),separate-uncertainty] S[table-format=3.2(2),separate-uncertainty]}
      \toprule
                                    &        & \multicolumn{2}{c}{\clap{\textbf{Long}}}                                       \\
      \cmidrule(lr){3-4} \cmidrule(lr){5-6}
      \textbf{Model}                & \textbf{Size}              & {Fixed Length}  & {Variable Length} \\
\midrule
      \LLaMmleinVec                 & 120M              & 6.69 \pm 0.14   & 8.39 \pm 0.35     \\
      \LLaMmleinVec                 & 1B     & 42.70 \pm 0.12  & 59.70 \pm 0.30    \\
      \LLaMmleinVec                 & 7B     & 180.00 \pm 0.19 & 304.00 \pm 0.41   \\
      \midrule
      \modernGBERT                  & 134M   & 5.42+-0.33      & 4.71+-0.75        \\
      \modernGBERT                  & 1B     & 28.70+-0.31     & 26.20+-0.36       \\
      \bottomrule
   \end{tabular}
   }
   \caption{Model throughput in seconds per million tokens. All models run
      on RTX A6000 with Bfloat16 and Flash Attention 2.
      Reported uncertainty is the empirical standard deviation on 10 repetitions.}\label{tab:efficiency-small}
\end{table}

We evaluate inference efficiency across varying sequence lengths using four synthetic datasets, each containing \num{8192} documents composed of random tokens.
Following \citet{warner2024smarterbetterfasterlonger}, two datasets use fixed-length sequences
(\num{512} and \num{8192} tokens), while the other two sample sequence lengths from normal distributions (either mean 256, variance 64; or mean \num{4096}, variance \num{1024}) to better simulate real-world conditions.
Our \modernGBERT models adopt ModernBERT's unpadding approach:
padding tokens are removed and sequences in a batch are concatenated, allowing Flash Attention to handle variable-length attention masks.
The computational equivalence is facilitated by carefully crafting an appropriate attention mask.
All other models rely on conventional padding.
Among smaller models (134M resp.\@ 120M), \modernGBERT and \LLaMmleinVec achieve comparable efficiency on fixed-length data, both only surpassed by \GbertBASE{} and \XLMRobertaBASE{} in terms of efficiency on short sequences.
For 1B variants, \modernGBERT consistently outperforms \LLaMmleinVec 1B and 7B variations in inference speed, likely due to its architectural decisions optimized for efficiency, such as ensuring that weight matrices have dimension of multiples of 64, and are divisible
into 128 $\times$ 256 block for efficient tiling on the GPU.
Gains are most pronounced for variable-length datasets, where \modernGBERT's unpadding yields clear benefits (see \Cref{tab:efficiency-small}):
the 134M \modernGBERT is the most efficient model on variable length, and the 1B variant substantially outpaces its \LLaMmleinVec counterpart.
Notably, \modernGBERT 1B matches the MTEB performance of \LLaMmleinVec 7B (see \Cref{tab:mteb_small}) while being ten times faster on long-context documents.
\cbox{Observation}{When considering the trade-off between computational efficiency and downstream performance metrics, \modernGBERT consistently emerges as the optimal solution—frequently outperforming \LLaMmleinVec on both dimensions simultaneously.}

\section{Related Work}
\paragraph{Next-Generation Encoders}
Several recent efforts have extended \ModernBERT to other languages and domains, such as French \citep{antoun2024camembert20smarterfrench} and Japanese \citep{sugiura2025llmjpmodernbertmodernbertmodeltrained}.
Recently, \mmBERT was created - a \ModernBERT model trained on 3T tokens of multilingual text covering over \num{1800} languages \citep{marone2025mmbertmodernmultilingualencoder}.

Concurrent to our work, several alternative encoder architectures have been proposed.
\citet{breton_neobert_2025} introduced NeoBERT, a 250M parameter English encoder incorporating similar architectural innovations like \ModernBERT, but scaling up layers rather than hidden dimension, switching from GeLU to SwiGLU activation, and using a modified training scheme (Cosine scheduler, reduced masking). Their model surpasses \ModernBERT-large on GLUE and MTEB with 100M fewer parameters, although its scalability with model size remains unexplored.
Likewise, \citet{boizard2025eurobert} recently presented \euroBERT (210M, 610M, 2.1B), a multilingual encoder family featuring architectural changes similar to those of \ModernBERT, but retaining some architectural details (RMSNorm layer normalization, SiLU activation function, Llama-style tokenizer) from the \Llama family, resembling our \LLaMmleinVec architecture.

\citet{antoun2025modernbertdebertav3examiningarchitecture} compared French \ModernBERT and DeBERTaV3, finding DeBERTaV3 to be superior on downstream tasks but significantly slower in training and inference.

\paragraph{Tuning decoder-only LLMs into Encoders}
Due to the lack of new encoder models \citep{weller2025seqvsseqopen}, works like \LLMVec (\Cref{sec:methodology_llm2vec}) have proposed strategies for turning decoder-only models into encoders (\Cref{sec:methodology_llm2vec}).
Few works have investigated converting decoder-only LLMs into encoders, besides \LLMVec (\Cref{sec:methodology_llm2vec}).
Recent studies predominantly address either distilling text embedders \citep{li-li-2024-bellm,lee2025nvembedimprovedtechniquestraining,lee2024geckoversatiletextembeddings,ma2025dramadiverseaugmentationlarge} or fine-tuning LLMs as bidirectional encoders for specific tasks \citep{li2023labelsupervisedllamafinetuning,dukic-snajder-2024-looking}, with evaluation primarily focused on English or multilingual settings.

\citet{khosla2025magnetaugmentinggenerativedecoders} introduced MAGNET, a method for converting decoder-only models into foundational encoders, similarly to \LLMVec.
Unlike \LLMVec, which enables bidirectional attention and masked next-token prediction, MAGNET uses a hybrid of bidirectional and causal attention and adds a missing-span generation objective, aiming for a more general pre-training signal.
Concurrent to our work, the Ettin Suite \citep{weller2025seqvsseqopen} provides a systematic comparison of encoder-only and decoder-only architectures trained on identical data and recipes.
Notably, they also experiment with objective switching, continuing training on 50B tokens, and directly comparing downstream task performance.

\section{Conclusion}
In this work we introduce two encoder-only families from scratch \modernGBERT and \LLaMmleinVec, contributing:
\begin{enumerate*}[label=\alph*)]
    \item the first systematic MLM vs.\ \LLMVec comparison for encoders using identical datasets,
    \item a novel \QANIAH evaluation adapted for encoder models, and
    \item the first combination of \LLMVec with context extension training.
\end{enumerate*}

Our main findings show that the proposed \modernGBERT family, especially the 1B variant, sets a new state-of-the-art for German encoders, outperforming previous models while remaining suitable for practical deployment as a drop-in replacement for GBERT, capable of handling sequences of up to \num{8192} tokens.
Our learning dynamics analysis confirms that larger encoder architectures can effectively exploit terabyte-scale German monolingual corpora, with performance consistently improving with increased model size and data.
A comparison of \modernGBERT and \LLaMmleinVec (derived from \LLaMmlein) shows that dedicated encoder training yields superior results, justifying its computational expense when parameter efficiency is essential.
At the same time, \LLMVec provides a resource-efficient alternative for scenarios where training a full encoder from scratch is too expensive.
Notably, in some use cases, we find that fine-tuning a decoder with a bidirectional attention mask already delivers substantial gains.
These trends suggest that even larger encoder models could yield further gains, which we leave to future work.
By releasing \modernGBERT and \LLaMmleinVec, along with full training transparency, intermediate checkpoints, detailed documentation, and accompanying resources, we aim to facilitate further development and understanding within the German LLM community.

\newpage
\section*{Limitations}
Despite the \modernGBERT and \LLaMmleinVec models being a notable advancement in the German NLP landscape, several limitations persist:
\begin{enumerate*}[label=\textbf{\arabic*)}]
    \item \textbf{Monolingual focus.} Although the focus on German is a strength for this specific context, \modernGBERT is unable to utilize multilingual contexts or perform cross-lingual tasks, hindering applicability in some scenarios.
    \item \textbf{Limited coding capabilities.} High-quality German resources for coding are rare, and no code is included in the training dataset. This restricts its capabilities in code retrieval applications.
    \item \textbf{Evaluation scope.} While we rigorously evaluated our models on the German SuperGLEBer and MTEB benchmarks, these benchmarks are limited in their domain, and other domains such as literature, medical domains, or technical subjects were not tested. Furthermore, our benchmarks do not strictly probe for ``German factual knowledge'', for instance, knowledge about German geography, or common German TV shows.
    \item \textbf{No custom tokenizer.} We utilized the original BERT-style GBERT tokenizer due to its availability and persistent usage. However, we did not invest in developing a custom tokenizer, like the BPE-style OLMo tokenizer used in ModernBERT. Consequently, \modernGBERT's tokenizer cannot, e.g., differentiate between various whitespace characters or encode emoji.
    \item \textbf{Evaluation of long-context understanding.} Due to the absence of high-quality native German evaluation datasets, we had to rely on non-natural QA-NIAH sequences, only broadly testing for long-context understanding. Contrast this with English benchmarks such as $\infty$Bench-MC \citep{zhang-etal-2024-bench} or LongBench-v2 \citep{bai2025longbenchv2deeperunderstanding}, which include full novels along with questions that require attention to much information scattered throughout the novel. In future work, we plan on developing a dedicated high-quality non-synthetic German long-context evaluation benchmark.
\end{enumerate*}

\section{Bibliographical References}\label{sec:reference}

\bibliographystyle{lrec2026-natbib}
\bibliography{anthology,custom}

\end{document}